\documentclass{article}
\usepackage{spconf,amsmath,graphicx}
\usepackage{comment}

\usepackage{times}
\usepackage{microtype}
\usepackage{epsfig}
\usepackage[table,xcdraw]{xcolor}
\usepackage{caption}
\usepackage{float}
\usepackage{placeins}
\usepackage{color, colortbl}
\usepackage{stfloats}
\usepackage{enumitem}
\usepackage{tabularx}
\usepackage{xstring}
\usepackage{multirow}
\usepackage{xspace}
\usepackage{url}
\usepackage{subcaption}
\usepackage{xcolor}
\usepackage[hang,flushmargin]{footmisc}
\usepackage[cjk]{kotex}
\usepackage[normalem]{ulem}
\usepackage{multirow}
\usepackage{booktabs}
\usepackage{caption}
\usepackage{vcell}
\usepackage{lipsum}
\usepackage[utf8]{inputenc}
\usepackage{amsmath}
\usepackage{verbatim}
\usepackage{color} 
\usepackage{textcomp}
\usepackage{algorithm} 
\usepackage{algpseudocode} 
\usepackage{multirow}
\usepackage[normalem]{ulem}
\useunder{\uline}{\ul}{}
\usepackage{bbding}
\usepackage{pifont}
\usepackage{wasysym}
\usepackage{amssymb}

\title{Towards Robust Multimodal Prompting with Missing Modalities}
\name{Jaehyuk Jang, Yooseung Wang, Changick Kim\thanks{We designed figures using images from Flaticon.com in this paper.}}
\address{School of Electrical Engineering, Korea Advanced Institute of Science and Technology, South Korea}

\begin{document}
%
\maketitle
\begin{abstract}

Recently, multimodal prompting, which introduces learnable missing-aware prompts for all missing modality cases, has exhibited impressive performance. 
However, it encounters two critical issues: 1) The number of prompts grows exponentially as the number of modalities increases; and 2) It lacks robustness in scenarios with different missing modality settings between training and inference.
In this paper, we propose a simple yet effective prompt design to address these challenges.
Instead of using missing-aware prompts, we utilize prompts as modality-specific tokens, enabling them to capture the unique characteristics of each modality.
%
%
%
Furthermore, our prompt design leverages orthogonality between prompts as a key element to learn distinct information across different modalities and promote diversity in the learned representations.
Extensive experiments demonstrate that our prompt design enhances both performance and robustness while reducing the number of prompts.

\end{abstract}

\begin{keywords}
Multimodal, Missing modality, Prompt learning
\end{keywords}
\vspace{-0.2cm}
\section{Introduction}
\label{sec:intro}

Advances in sensing technology allow us to acquire a variety of signals including images, text, heat map, depth map, and more.
Various methods have emerged for multimodal downstream tasks, such as classification~\cite{clf1}, action recognition~\cite{li2023av, chen2022mm, ar1, ar2}, and emotion recognition~\cite{huang2020multimodal, dutta2022multimodal, yang2022disentangled}.
Recently, several methods~\cite{clf1, woo2023towards} have been proposed based on transformer~\cite{transformer, vit} to deal with the missing modality scenario during the inference phase.
Nevertheless, these works assume that the training dataset is full-modality.
In real-world scenarios, users may acquire modality-incomplete data due to device constraints regardless of the training and inference phase.
Moreover, expanding the number of modalities results in high computational and memory costs, making it more challenging to train the entire model.

Inspired by prompt learning~\cite{jia2022visual, wang2022learning, pl1, pl2}, Lee \textit{et al.}~\cite{lee2023multimodal} propose multimodal prompting,
which adopts learnable tokens, missing-aware prompts (MAPs), to transfer knowledge from pretrained models on large-scale datasets to downstream tasks.
This method focuses solely on training MAPs and parameters associated with downstream tasks while keeping the backbone network frozen.
%
%
By assigning prompts to missing modality cases, prompts are updated while learning how to deal with each missing modality case. 
As a result, it can handle general scenarios where modality-incomplete data exist during the training phase, requiring less than 1$\%$ of the total parameters to finetune the pretrained model.

\begin{figure}[t]
   \centering
   \includegraphics[width=\linewidth]{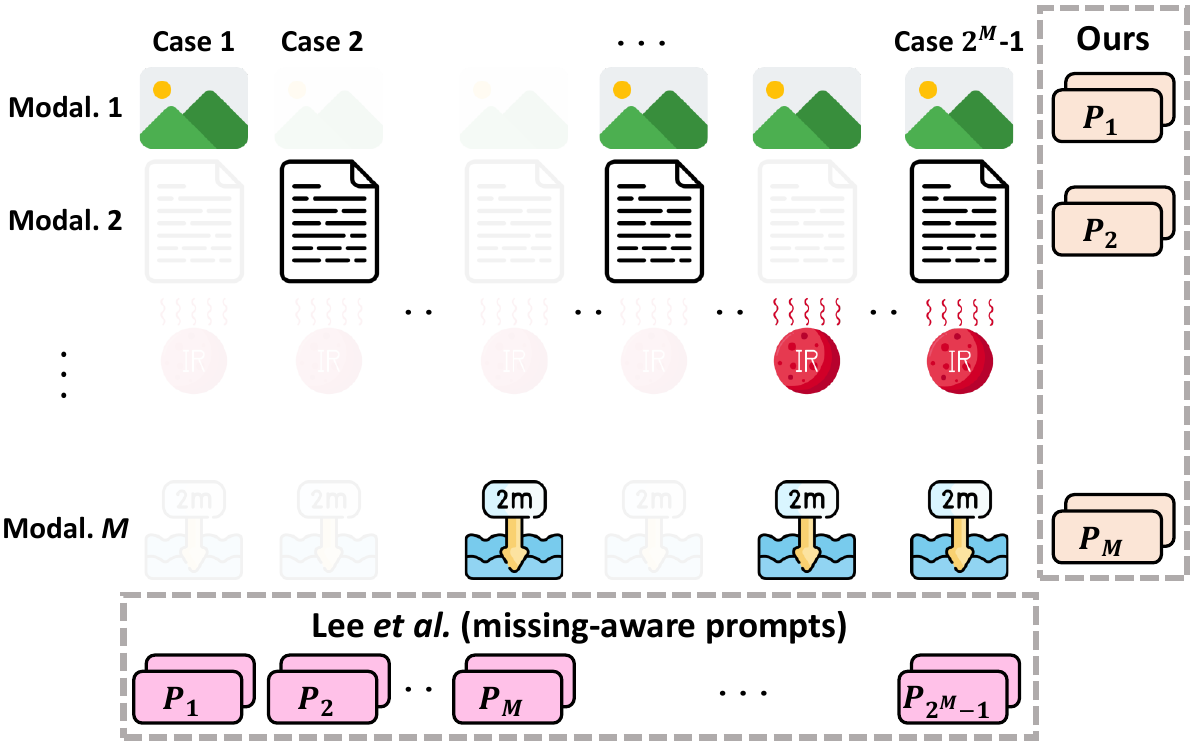}
   \vspace{-0.5cm}
   \caption{
   %
    %
    Illustrating our motivation, consider a scenario with a total of $M$ modalities.
    In the previous work~\cite{lee2023multimodal}, addressing all missing modality cases requires $2^M-1$ missing-aware prompts.
    In contrast, our approach involves only allocating $M$ prompts, each designed to capture the unique characteristics of each individual modality.
   }
   \label{fig:motivation}
    \vspace{-0.6cm}
\end{figure}

%
%
While MAPs are a promising solution for addressing the missing modality scenario during both training and inference, there are two notable challenges.
First, the number of prompts exhibits exponential growth as the number of modalities increases.
%
%
For a $M$ modality task, the requisite number of MAPs is $2^M-1$ to treat all missing modality scenarios, as depicted in Figure~\ref{fig:motivation}.
This predicament is critical as the advancement of artificial intelligence ensures the inexorable integration of an increasing number of modalities~\cite{imagebind}.
%
Second, it lacks robustness in unseen missing modality settings at the training phase.
Unless an ample number of samples are available for all missing modality cases, the model remains incapable of addressing unseen missing modality cases that were not encountered during the training phase.
For example, if we can leverage only modality-complete cases and text-only cases in a training phase, the model can not deal with image-only cases in an inference phase.

In this paper, we propose a simple yet effective prompt design to treat these issues.
%
%
%
%
Note that we are focusing on the image-text multimodal classification task.
In contrast to the previous study~\cite{lee2023multimodal}, which creates MAPs for each missing modality case (image-only, text-only, and modality-complete), we are developing modality-specific prompts (MSPs) for each individual modality.
%
%
%
%
%
%
MSPs acquire knowledge specific to modality from the backbone pretrained on large-scale datasets.
We utilize image-specific prompt and text-specific prompt as input of multimodal transformer for image-only and text-only cases, respectively.
We adopt a combination of image-specific prompt and text-specific prompt for modality-complete cases so that both prompts can be updated simultaneously.
In this way, we train the prompts assigned to the modalities present in samples, disregarding the missing modality case.
Furthermore, we incorporate an additional loss term to enforce orthogonality among MSPs, ensuring the diversity of information learned across different modalities.
Extensive experiments show that the proposed straightforward prompt design boosts the performance and significantly enhances the robustness for unseen missing modality settings in the training phase.
We expect that our prompt design will serve as a valuable guiding principle for a wider range of modality tasks, paving the way for further advancements in this field.

\begin{figure*}[t!]
  \centering
  \includegraphics[width=\linewidth]{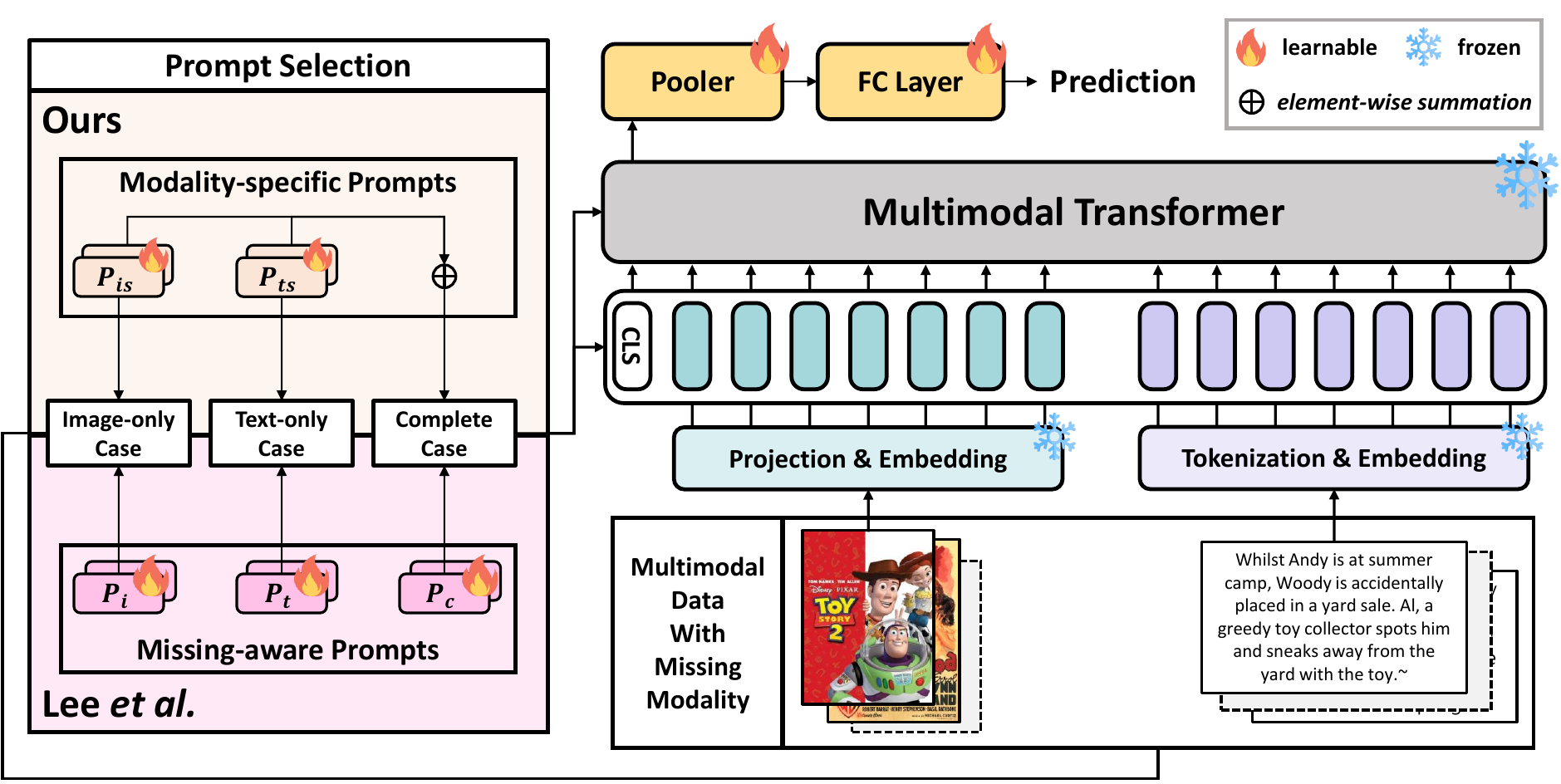}
   \caption{Illustration of our modality-specific prompts design and overall framework.}
   \label{fig:framework}
   \vspace{-0.5cm}
\end{figure*}

\vspace{-0.3cm}
\section{Method}
\label{sec:method}
\vspace{-0.2cm}

\subsection{Problem Definition} 
We utilize a multimodal dataset labeled as $D$, comprising two types of modality (i.e., text and image) denoted as $m_1$ and $m_2$. The multimodal dataset is symbolized as $D=\{D_c, D_{m_1}, D_{m_2}\}$. Here, $D_c = \{x_{m_1}^i, x_{m_2}^i, y^i\}$ signifies the complete set containing both modality input and label, while $D_{m_1} = \{x_{m_1}^j, y^j\}$ and $D_{m_2} = \{x_{m_2}^k, y^k\}$ represent subsets with a single modality. In accordance with prior work~\cite{lee2023multimodal}, we assign meaningless values to missing-modality instances, like empty text or an image with a pixel value of one. To retain the multimodal input structure, we introduce placeholder inputs, $\tilde{x}_{m_1}$ and $\tilde{x}_{m_2}$ (e.g., empty strings or pixels for text/images), for the absent modality data. This leads to $\tilde{D}_{m_1} = \{x_{m_1}^j, \tilde{x}_{m_2}^j, y^j\}$ and $\tilde{D}_{m_2} = \{\tilde{x}_{m_1}^k, x_{m_2}^k, y^k\}$. Consequently, the multimodal dataset with missing modalities is denoted as $\tilde{D} = \{D_c, \tilde{D}_{m_1}, \tilde{D}_{m_2}\}$. This results in datasets covering different scenarios of missing data, including complete data $D_c$, text-only data $D_{m_1}$, and image-only data $D_{m_2}$. The overall dataset utilized in the experiments is represented as $\tilde{D}$. The objective of the task is to enhance the ability to generalize in cases where modalities are missing. To tackle this, we concentrate on MAPs, as proposed in the previous study~\cite{lee2023multimodal}, to strengthen the resilience of prompts when dealing with missing modalities.

\vspace{-0.2cm}
\subsection{Revisiting multimodal prompting}
\vspace{-0.2cm}

For the image-text multi-modality, there are three missing modality cases (i.e., image-only, text-only, and modality-complete).
Lee \textit{et al.}~\cite{lee2023multimodal} introduce MAPs to categorize input data.
%
Depending on the missing modality case of the input data, the corresponding prompt is selected.
For example, $P_i$ for the image-only sample, $P_t$ for the text-only sample, and $P_c$ for the modality-complete sample as depicted in Fig.~\ref{fig:framework}.
%
%
The prompt is concatenated to the feature in the multimodal transformer and updated through backpropagation in the training phase.
%
%
In this way, the prompt distinguishes between all missing modality cases, and the pretrained backbone transfers knowledge to the prompt accordingly.

\vspace{-0.2cm}
\subsection{Prompt design}
\vspace{-0.2cm}
Instead of using prompts for missing awareness, we design MSPs that capture the distinct characteristics of each modality.
Note that $P_{is}$ and $P_{ts}$ are our image-specific prompts and text-specific prompts, respectively.
Each prompt learns knowledge from the pretrained backbone specific to its corresponding modality.
As shown in Fig.~\ref{fig:framework}, we select $P_{is}$ and $P_{ts}$ for the image-only case and text-only case, respectively.
Instead of $P_c$ in MAPs, we do not have a modality-complete case-specific prompt.
Rather, we simply combine $P_{is}$ and $P_{ts}$ to handle the modality-complete case.
By utilizing the element-wise summation of two prompts, both prompts $P_{is}$ and $P_{ts}$ are updated when the input is a modality-complete case.
%
%
Generating differentiable complete case-specific prompts based on the given text and image modalities, rather than random generalization as used in MAPs, allows each prompt to learn more frequently while reducing the number of prompts. 
%

Furthermore, we incorporate an additional loss term $L_{ortho}$ to enforce orthogonality among MSPs.
%
By reducing the similarity between $P_{is}$ and $P_{ts}$, we maximize the gap between the two prompts to ensure the diversity of information learned from image and text.
Therefore, we minimize the cosine similarity between $P_{ts}$ and $P_{is}$ as follows:  

\begin{equation}
    L_{ortho} = \frac{|f(P_{is}) \cdot f(P_{ts})|}{\max \left(\left\|f(P_{is})\right\|_2 \cdot\left\|f(P_{ts})\right\|_2, \epsilon\right)},
    \label{eq:L_ortho}
\end{equation}
where $f$ is the flatten function, $|.|$ is the absolute value function, $\left\|.\right\|_2 $ is the L2-norm, and $\epsilon$ is a small value to avoid division by zero.

\vspace{-0.2cm}

\vspace{-0.3cm}
\subsection{Training objectives}
\vspace{-0.2cm}
For finetuning, we freeze all parameters in the text encoder, image encoder, and multimodal transformer and train only MSPs and task-specific layers (pooler and classifier).
We note that $L_{cls}$ is a task-specific objective function (e.g., binary cross-entropy loss for movie genre classification and cross-entropy loss for food classification).
Finally, the overall objective function $L_{total}$ for finetuning our model can be represented as follows:
\begin{equation}
    L_{total} = L_{cls} + \lambda L_{ortho},
    \label{eq:L_total}
\end{equation}
where $\lambda$ is the balancing hyperparameter.

\vspace{-0.3cm}
\section{Experiments}
\label{sec:experiment}
\vspace{-0.2cm}

\subsection{Datasets and evaluation metrics}

MM-IMDb~\cite{mmimdb} is a comprehensive movie genre multi-label classification dataset that incorporates both image and text modalities.
The dataset offers various multi-label movie categories, such as action, comedy, drama, thriller, and more.
UPMC Food-101~\cite{food101} is a food category multi-class classification dataset including image and text. All images are collected from the Google Image Search engine and not post-filtered by human labor so that each category contains about $5\%$ noisy images. 
In the case of MM-IMDb, the multi-label classification performance is evaluated using F1-Macro. For UPMC Food-101, we employ classification accuracy as the evaluation metric.

\vspace{-0.2cm}
\subsection{Implementation details}
\vspace{-0.2cm}
For a fair comparison, we followed the experiment settings with previous work~\cite{lee2023multimodal}.
We adopted a multimodal transformer ViLT~\cite{kim2021vilt} pretrained on large-scale vision-language datasets as our backbone network. 
For MAPs and our MSPs, we used input-level prompting, which attaches prompts to the input of the multi-head self-attention layer.
We finetuned the entire pretrained ViLT with a target dataset for comparison. 
In contrast, we froze the pretrained ViLT and solely trained prompts and task-specific layers for MAPs and our MSPs as depicted in Fig~\ref{fig:framework}.
We set the missing rate to 70\% in all our experiments, which is the total percentage of modalities removed among image or text modalities. 
To consider various missing modality scenarios, we set up a total of three scenarios, including two scenarios in which 70\% of image or text were removed and a scenario in which both modalities were equally missing by 35\%. 
Three scenarios were used for both training and inference as shown in Table~\ref{table:result} and ~\ref{tab:ablation}. 
We set the balancing hyperparameter $\lambda$ to 0.15 and 0.1 for MM-IMDb and UPMC Food-101 datasets, respectively.
We conducted all experiments, including ViLT, MAPs, and our method, with the same initial learning rate $1\times10^{-2}$, weight decay $2\times10^{-2}$, AdamW optimizer~\cite{AdamW}, and batch size of 6 by utilizing an NVIDIA RTX 3090 GPU.
%

\vspace{-0.2cm}
\subsection{Comparisons with the previous method}
\vspace{-0.2cm}
We show quantitative results of ViLT, MAPs, and our MSPs on MM-IMDb and UPMC Food-101 datasets in Table~\ref{table:result}. 
%
Compared to ViLT, MAPs demonstrate superior performance when the missing modality cases encountered during inference align with those seen during training.
However, MAPs can not cope with the unseen case where the modality composition ratio of the training and inference is unbalanced. 
For example, consider a training set consisting of 100$\%$ image-modality and 30$\%$ text-modality samples. 
In this scenario, MAPs successfully train the modality-complete prompt $P_c$ and the image-only prompt $P_i$ while the text-only prompt $P_t$ remains unlearned.
Consequently, the prediction of a text-only sample is likely inaccurate due to the absence of information in $P_t$.
Due to the reasons mentioned above, MAPs lack robustness when faced with missing modality cases that were not encountered during the training phase, and they even demonstrate inferior performance compared to ViLT.
%
%
%
On the other hand, it shows that proposed MSPs achieve robustly uniform performance improvement across various missing modality scenarios.
%
Regardless of the missing modality case, our approach trains prompt for learning information from each modality in the input sample, as mentioned in Section~\ref{sec:method}.
%
%
%
%
In summary, MSPs require fewer prompts than MAPs while outperforming both ViLT and MAPs in terms of both performance and robustness.



\begin{table*}[t!]
\centering
\resizebox{.94\textwidth}{!}{
\begin{tabular}{c|c|cc|cc|cccc}
\toprule
\multirow{2}{*}{Datasets}                                                     & \multirow{2}{*}{Missing Rate} & \multicolumn{2}{c|}{Training}                   & \multicolumn{2}{c|}{Inference} & \multirow{2}{*}{ViLT~\cite{kim2021vilt}} & \multirow{2}{*}{MAPs~\cite{lee2023multimodal}} & \multicolumn{2}{c}{\multirow{2}{*}{\begin{tabular}[c]{@{}c@{}}MSPs\\ (Ours)\end{tabular}}} \\
                                                                              &                               & Image                  & Text                   & Image       & Text        &                       &                      &                                                                       \\ \hline
\multirow{9}{*}{\begin{tabular}[c]{@{}c@{}}MM-IMDb~\cite{mmimdb}\\ (F1-Macro)\end{tabular}} & \multirow{18}{*}{70\%}        & \multirow{3}{*}{100\%} & \multirow{3}{*}{30\%}  & 100\%       & 30\%        & 34.26                 & {\ul 36.89}          & \textbf{38.34}                                                        \\
                                                                              &                               &                        &                        & 30\%        & 100\%       & {\ul 27.25}           & 20.17                & \textbf{34.11}                                                        \\
                                                                              &                               &                        &                        & 65\%        & 65\%        & {\ul 32.51}           & 29.79                & \textbf{36.97}                                                        \\ \cline{3-9} 
                                                                              &                               & \multirow{3}{*}{30\%}  & \multirow{3}{*}{100\%} & 100\%       & 30\%        & {\ul 30.00}           & 24.31                & \textbf{34.32}                                                        \\
                                                                              &                               &                        &                        & 30\%        & 100\%       & 37.63                 & {\ul 47.18}          & \textbf{47.45}                                                        \\
                                                                              &                               &                        &                        & 65\%        & 65\%        & 34.47                 & {\ul 37.25}          & \textbf{42.17}                                                        \\ \cline{3-9} 
                                                                              &                               & \multirow{3}{*}{65\%}  & \multirow{3}{*}{65\%}  & 100\%       & 30\%        & 33.76                 & \textbf{37.80}       & {\ul 37.30}                                                           \\
                                                                              &                               &                        &                        & 30\%        & 100\%       & 36.48                 & {\ul 42.72}          & \textbf{44.81}                                                        \\
                                                                              &                               &                        &                        & 65\%        & 65\%        & 36.54                 & {\ul 40.84}          & \textbf{42.03}                                                        \\ \cline{1-1} \cline{3-9} 
\multirow{9}{*}{\begin{tabular}[c]{@{}c@{}}Food-101~\cite{food101}\\ (Accuracy)\end{tabular}} &                               & \multirow{3}{*}{100\%} & \multirow{3}{*}{30\%}  & 100\%       & 30\%        & 66.01                 & {\ul 73.71}          & \textbf{73.77}                                                        \\
                                                                              &                               &                        &                        & 30\%        & 100\%       & {\ul 43.67}           & 27.46                & \textbf{57.22}                                                        \\
                                                                              &                               &                        &                        & 65\%        & 65\%        & {\ul 54.75}           & 50.79                & \textbf{65.13}                                                        \\ \cline{3-9} 
                                                                              &                               & \multirow{3}{*}{30\%}  & \multirow{3}{*}{100\%} & 100\%       & 30\%        & \uline{41.94}                 & 29.71                & \textbf{56.87}                                                        \\
                                                                              &                               &                        &                        & 30\%        & 100\%       & {\ul 76.40}                 & \textbf{86.34}       &  \textbf{86.34}                                                           \\
                                                                              &                               &                        &                        & 65\%        & 65\%        & {\ul 59.03}           & 57.92                & \textbf{71.88}                                                        \\ \cline{3-9} 
                                                                              &                               & \multirow{3}{*}{65\%}  & \multirow{3}{*}{65\%}  & 100\%       & 30\%        & 64.25                 & {\ul 71.17}          & \textbf{71.58}                                                        \\
                                                                              &                               &                        &                        & 30\%        & 100\%       & 73.62                 & {\ul 85.55}          & \textbf{85.91}                                                        \\
                                                                              &                               &                        &                        & 65\%        & 65\%        & 68.60                 & {\ul 78.49}          & \textbf{78.89}                                                       
\\
\bottomrule
\end{tabular}
}
\vspace{-0.19cm}
\caption{Comparison on the MM-IMDb~\cite{mmimdb} and UPMC Food-101~\cite{food101} datasets with missing rate=70$\%$ under various training and inference missing-modality cases. 
We reproduced previous work using official code with our experiment setting, so the results of previous work are different from the original paper.
The best value is in \textbf{bold}, and the second best is \uline{underlined}.
}
\label{table:result}
\vspace{-0.4cm}
\end{table*}

\begin{table}[t]
\centering
\resizebox{\columnwidth}{!}{
\begin{tabular}{cc|cc|ccc}
\toprule
\multicolumn{2}{c|}{Training}                   & \multicolumn{2}{c|}{Inference} & \multirow{2}{*}{ViLT~\cite{kim2021vilt}} & \multirow{2}{*}{+MSPs} & \multirow{2}{*}{+$L_{ortho}$} \\
Image                  & Text                   & Image        & Text       &                       &                       &                        \\ \hline
\multirow{3}{*}{100\%} & \multirow{3}{*}{30\%}  & 100\%        & 30\%       & 34.26                 & {\ul 37.83}           & \textbf{38.34}         \\
                       &                        & 30\%         & 100\%       & 27.25                 & {\ul 32.11}           & \textbf{34.11}         \\
                       &                        & 65\%         & 65\%       & 32.51                 & {\ul 34.90}            & \textbf{36.97}         \\ \hline
\multirow{3}{*}{30\%}  & \multirow{3}{*}{100\%} & 100\%        & 30\%       & 30.00                    & {\ul 33.22}           & \textbf{34.32}         \\
                       &                        & 30\%         & 100\%       & 37.63                 & {\ul 46.95}           & \textbf{47.45}         \\
                       &                        & 65\%         & 65\%       & 34.47                 & {\ul 41.24}           & \textbf{42.17}         \\ \hline
\multirow{3}{*}{65\%}  & \multirow{3}{*}{65\%}  & 100\%        & 30\%       & 33.76                 & \textbf{38.07}        & {\ul 37.30}             \\
                       &                        & 30\%         & 100\%       & 36.48                 & {\ul 44.14}           & \textbf{44.81}         \\
                       &                        & 65\%         & 65\%       & 36.54                 & {\ul 41.38}           & \textbf{42.03}
\\
\bottomrule
\end{tabular}
}
\vspace{-0.25cm}
\caption{Ablation study of our modality-specific prompts and orthogonal loss $L_{ortho}$ on MM-IMDb~\cite{mmimdb} with missing rate=70$\%$. The evaluation metric is F1-Macro.
The best value is in \textbf{bold}, and the second best is \uline{underlined}.
}
\label{tab:ablation}
\vspace{-0.5cm}
\end{table}

\subsection{Ablation study}
As shown in Table~\ref{tab:ablation}, we conducted an ablation study on MM-IMDb to verify that our MSPs and orthogonal loss $L_{ortho}$ are effective.
MSPs not only yield improvements in performance but also enhance the robustness for unseen missing modality cases. 
%
Furthermore, $L_{ortho}$ guides the prompts $P_{ts}$ and $P_{is}$ to be orthogonal and acquire distinct information from each modality.
This strategy significantly contributes to enhancing performance.
%
In conclusion, our study demonstrates that leveraging our novel prompt design and orthogonal loss for extracting unique information from each modality improves both performance and robustness.

\vspace{-0.2cm}
\section{Conclusion}
\label{sec:conclusion}
\vspace{-0.4cm}

In this paper, we have proposed a simple yet effective prompt design for robust real-world missing modality scenarios.
We introduce modality-specific prompts that learn distinct information from each modality distribution so that they are missing-modality-case-agnostic.
%
Prompt-based learning with orthogonal loss function enforces orthogonality among prompts and enlarges the diversities between different modalities. 
%
Evaluations conducted on two benchmark datasets show that our method outperforms in terms of performance and robustness compared to previous work.
As the integration of increasingly diverse multimodal signals, we firmly believe that our prompt design will open the way for multimodal finetuning with missing modalities.
In the future, we will apply our technique to the task with more modality settings, e.g., action recognition with RGB, depth, and IR as inputs.

\section{Acknowledgement}
\vspace{-0.3cm}
This work was supported by Institute of Information communications Technology Planning Evaluation (IITP) grant funded by the Korea government(MSIT) (No. 2021-0-00193, Development of photorealistic digital human creation and 30fps realistic rendering technology).

\bibliographystyle{elsarticle-num}

\newpage

\bibliography{5.reference}

\end{document}